\documentclass[sigconf]{acmart}

\usepackage{balance}
\usepackage{amsfonts}
\usepackage{booktabs}
\usepackage{dsfont}
\usepackage{bm}
\usepackage{amsmath}
\usepackage{subcaption}
\usepackage{algpseudocode}
\usepackage{algorithm}
\usepackage{microtype}
\usepackage{hyperref}
\usepackage{graphicx} 
\usepackage{caption}
\usepackage{graphicx}  
\usepackage{xcolor}
\newcommand{\para}[1]{\vskip 2mm\noindent\textbf{#1}~}

\newcommand{\spara}[1]{\vskip 1mm\noindent\emph{#1}:}

\DeclareRobustCommand{\sameer}[1]{{\sethlcolor{yellow}\hl{\textbf{sameer:} #1}}}

\def\BibTeX{{\rm B\kern-.05em{\sc i\kern-.025em b}\kern-.08emT\kern-.1667em\lower.7ex\hbox{E}\kern-.125emX}}
\newcommand{\hide}[1]{}

\DeclareMathOperator*{\argmin}{arg\,min}
\newcommand{\scaffolding}{scaffolding}

\def\BibTeX{{\rm B\kern-.05em{\sc i\kern-.025em b}\kern-.08emT\kern-.1667em\lower.7ex\hbox{E}\kern-.125emX}}
    
\copyrightyear{2020}
\acmYear{2020}
\setcopyright{acmcopyright}
\acmConference[AIES '20]{Proceedings of the 2020 AAAI/ACM Conference on AI, Ethics, and Society}{February 7--8, 2020}{New York, NY, USA}
\acmBooktitle{Proceedings of the 2020 AAAI/ACM Conference on AI, Ethics, and Society (AIES '20), February 7--8, 2020, New York, NY, USA}
\acmPrice{15.00}
\acmDOI{10.1145/3375627.3375830}
\acmISBN{978-1-4503-7110-0/20/02}

\begin{document}

\fancyhead{}

\title{Fooling LIME and SHAP: Adversarial Attacks on Post hoc Explanation Methods}

\author{Dylan Slack*}
\affiliation{%
  University of California, Irvine
}
\email{dslack@uci.edu}

\author{Sophie Hilgard*}
\thanks{*Both authors contributed equally to this research.  \\Code can be found at: \url{https://github.com/dylan-slack/Fooling-LIME-SHAP}}
\affiliation{%
  Harvard University}
\email{ash798@g.harvard.edu}

\author{Emily Jia}
\affiliation{%
  Harvard University
}
\email{ejia@college.harvard.edu}

\author{Sameer Singh}
\affiliation{University of California, Irvine}
\email{sameer@uci.edu}
 
\author{Himabindu Lakkaraju}
\affiliation{Harvard University}
\email{hlakkaraju@seas.harvard.edu}

%
\renewcommand{\shortauthors}{Slack and Hilgard, et al.}

%
\begin{abstract}
As machine learning black boxes are increasingly being deployed in domains such as healthcare and criminal justice, there is growing emphasis on building tools and techniques for explaining these black boxes in an interpretable manner. Such explanations are being leveraged by domain experts to diagnose systematic errors and underlying biases of black boxes. In this paper, we demonstrate that post hoc explanations techniques that rely on input perturbations, such as LIME and SHAP, are not reliable. Specifically, we propose a novel \emph{\scaffolding} technique that effectively hides the biases of any given classifier by allowing an adversarial entity to craft an arbitrary desired explanation. 
Our approach can be used to scaffold any biased classifier in such a way that its predictions on the input data distribution still remain biased, but the post hoc explanations of the scaffolded classifier look innocuous. 
Using extensive evaluation with multiple real world datasets (including COMPAS), we demonstrate how extremely biased (racist) classifiers crafted by our framework can easily fool popular explanation techniques such as LIME and SHAP into generating innocuous explanations which do not reflect the underlying biases. 
\end{abstract}

%
%

\ccsdesc[500]{Computing methodologies~Machine learning}
\ccsdesc[500]{Computing methodologies~Supervised learning by classification}
\ccsdesc[300]{Human-centered computing~Interactive systems and tools}

%
\keywords{black box explanations, model interpretability, bias detection, adversarial attacks}

\maketitle
\section{Introduction} \label{sec:intro}
Owing to the success of machine learning (ML) models, 
there has been an increasing interest in leveraging these models to aid decision makers (e.g., doctors, judges) in critical domains such as healthcare and criminal justice. The successful adoption of these models in domain-specific applications relies heavily on how well decision makers are able to understand and trust their functionality \cite{doshi2017towards,lipton2016mythos}. Only if decision makers have a clear understanding of the model behavior, can they diagnose errors and potential biases in these models, and decide when and how much to rely on them. However, the proprietary nature and increasing complexity of machine learning models makes it challenging for domain experts to understand these complex \emph{black boxes}, thus, motivating the need for tools that can explain them in a faithful and interpretable manner.

As a result, there has been a recent surge in post hoc techniques for explaining black box models in a human interpretable manner. 
One of the primary uses of such explanations is to help domain experts detect discriminatory biases in black box models ~\cite{tan2018distill,kim2017interpretability}. 
Most prominent of these techniques include \emph{local, model-agnostic} methods that focus on explaining individual predictions of a given black box classifier, including LIME~\cite{ribeiro16:kdd} and SHAP~\cite{lundberg17:a-unified}. 
These methods estimate the contribution of individual features towards a specific prediction by generating perturbations of a given instance in the data and observing the effect of these perturbations on the output of the black-box classifier. Due to their generality, these methods have been used to explain a number of classifiers, such as neural networks and complex ensemble models, and in various domains ranging from law, medicine, finance, and science~\cite{elshawi2019interpretability,ibrahim2019global,whitmore2016mapping}.
However, there has been little analysis of the reliability and robustness of these explanation techniques, especially in the adversarial setting, making their utility for critical applications unclear.

In this work, we demonstrate significant vulnerabilities in post hoc explanation techniques that can be exploited by an adversary to generate classifiers whose post hoc explanations can be arbitrarily controlled. 
More specifically, we develop a novel framework that can effectively mask the discriminatory biases of any black box classifier. 
Our approach exploits the fact that post hoc explanation techniques such as LIME and SHAP are perturbation-based, to create a \emph{\scaffolding} around any given biased black box classifier in such a way that its predictions on input data distribution remain biased, but its behavior on the perturbed data points is controlled to make the post hoc explanations look completely innocuous.
For instance, using our framework, we generate highly discriminatory scaffolded classifiers (such as the ones that \emph{only} use race to make their decisions) whose post hoc explanations (generated by LIME and SHAP) make them look completely innocuous, effectively hiding their discriminatory biases.

We evaluate the effectiveness of the proposed framework on multiple real world datasets --- COMPAS~\cite{larson2016we}, Communities and Crime~\cite{redmond2011communities}, and German loan lending~\cite{asuncion2007uci}. 
For each dataset, we craft classifiers that heavily discriminate based on protected attributes such as race (demographic parity ratio = 0), and show that our framework can effectively hide their biases.
In particular, our results show that the explanations of these classifiers generated using off-the-shelf implementations of LIME and SHAP do not flag \emph{any} of the relevant sensitive attributes (e.g., race) as important features of the classifier for any of the test instances, thus demonstrating that the adversarial classifiers successfully fooled these explanation methods. 
These results suggest that it is possible for malicious actors to craft adversarial classifiers that are highly discriminatory, but can effectively fool existing post hoc explanation techniques. This further establishes that existing post hoc explanation techniques are not sufficiently robust for ascertaining discriminatory behavior of classifiers in sensitive applications.

\section{Building Adversarial Classifiers to Fool Explanation Techniques}

In this section, we discuss our framework for constructing adversarial classifiers (\emph{scaffoldings}) that can fool post hoc explanation techniques which rely on input perturbations. 
We first provide a detailed overview of popular post hoc explanation techniques, namely, LIME~\cite{ribeiro16:kdd} and SHAP~\cite{lundberg17:a-unified}, and then present our framework for constructing adversarial classifiers. 

\subsection{Background: LIME and SHAP}
While simpler classes of models (such as linear models and decision trees) are often readily understood by humans, the same is not true for complex models (e.g., ensemble methods, deep neural networks). 
Such complex models are essentially black boxes for all practical purposes. 
One way to \emph{understanding} the behavior of such classifiers is to build simpler \emph{explanation models} that are interpretable approximations of these black boxes. 

To this end, several techniques have been proposed in existing literature. 
LIME~\cite{ribeiro16:kdd} and SHAP~\cite{lundberg17:a-unified} are two popular \emph{model-agnostic}, \emph{local explanation} approaches designed to explain any given black box classifier. These methods explain individual predictions of any classifier in an interpretable and faithful manner, by learning an interpretable
model (e.g., linear model) locally around each prediction.
%
Specifically, LIME and SHAP estimate feature attributions on individual instances, which capture the \emph{contribution} of each feature on the black box prediction. 
Below, we provide some details of these approaches, while also highlighting how they relate to each other.

Let $\mathcal{D}$ denote the input dataset of N data points i.e., $\mathcal{D} = (\mathcal{X},\bm{y}) = \{(x_1, y_1), (x_2, y_2) \cdots (x_N, y_N)\}$ where $x_i$ is a vector that captures the feature values of data point $i$, and $y_i$ is the corresponding class label. Let there be $M$ features in the dataset $\mathcal{D}$ and let $\mathcal{C}$ denote the set of class labels in $\mathcal{D}$ i.e., $y_i \in \mathcal{C}$. 
Let $f$ denote the black box classifier that takes a data point as input and returns a class label i.e.,  $f(x_i) \in \mathcal{C}$. The goal here is to explain $f$ in an interpretable and faithful manner. Note that neither LIME nor SHAP assume any knowledge about the internal workings of $f$.  Let $g$ denote an explanation model that we intend to learn to explain $f$. $g \in G$ where $G$ is the class of linear models. 

\hide{
\sameer{seems orthogonal to our contributions}
Often times, the features in the raw data (e.g., text, images) are not readily interpretable and therefore cannot be used in an explanation. To work around this problem, LIME and SHAP both assume that they have access to interpretable representations of orginal (raw) data points. More precisely, they assume access to a mapping function $h$ such that $x = h_{x}(x')$ where $x$ and $x' \in \{0,1\}^{M'}$ denote the original input data point and its corresponding \emph{simplified input} respectively, and $M'$ denotes the number of features in the new simplified input space. 
}

Let the complexity of the explanation $g$ be denoted as $\Omega(g)$ (complexity of a linear model can be measured as the number of non-zero weights), and let $\pi_{x}(x')$ denote the proximity measure between inputs $x$ and $x'$, to define the vicinity (neighborhood) around $x$. 
With all this notation in place, the objective function for both LIME and SHAP is crafted to generate an explanation that: (1) approximates the behavior of the black box accurately within the vicinity of $x$, and (2) achieves lower complexity and is thereby interpretable. 
\begin{equation}
    \argmin_{g \in \mathcal{G}} L(f, g, \pi_{x}) + \Omega(g)
\end{equation}
where the loss function $L$ is defined as: \[L(f, g, \pi_{x})  = \sum_{x' \in X'} [f(x') - g(x')]^2 \pi_{x}(x')\]
where $X'$ is the set of inputs constituting the neighborhood of $x$. 

The primary difference between LIME and SHAP lies in how $\Omega$ and $\pi_{x}$ are chosen. In LIME, these functions are defined heuristically: $\Omega(g)$ is the number of non-zero weights in the linear model
and $\pi_{x}(x')$ is defined using cosine or $l2$ distance. 
On the other hand, (Kernel) SHAP grounds these definitions in game theoretic principles to guarantee that the explanations satisfy certain desired properties.
More details about the intuition behind the definitions of these functions and their computation can be found in \citet{ribeiro16:kdd} and \citet{lundberg17:a-unified}.

\begin{figure}
    \centering
    \includegraphics[scale=0.5,clip,trim=0 0 0 40]{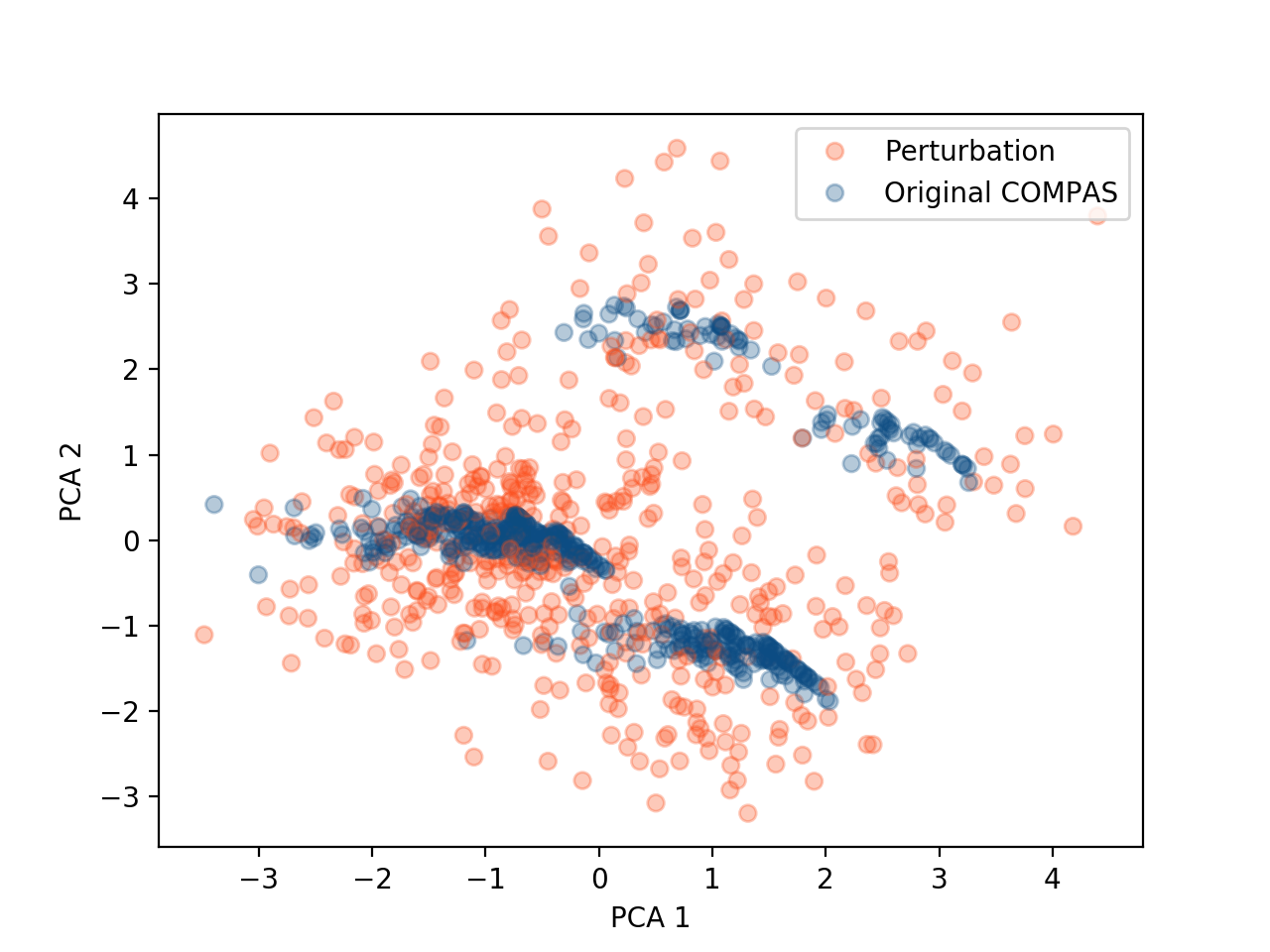}
    \caption{PCA applied to the COMPAS dataset (blue) as well as its LIME style perturbations (red).  Even in this low-dimensional space, we can see that data points generated via perturbations are distributed very differently from instances in the COMPAS data. In this paper, we exploit this difference to craft adversarial classifiers.}
    \label{fig:pca}
    \vspace{-0.2in}
\end{figure}

\subsection{Proposed Framework} 
In this section, we discuss our framework in detail. 
First, we discuss some preliminary details about our set up. Then, we discuss the intuition behind our approach. Lastly, we present the technical details of our approach along with a discussion of some of our design choices and implementation details.

\para{Preliminaries}
\spara{Setting} Assume that there is an adversary with an incentive to deploy a biased classifier $f$ for making a critical decision (e.g., parole, bail, credit) in the real world. The adversary must provide black box access to customers and regulators~\cite{regulation2016regulation}, who may use post hoc explanation techniques to better understand $f$ and determine if $f$ is ready to be used in the real world. If customers and regulators detect that $f$ is biased, they are not likely to approve it for deployment.  The goal of the adversary is to fool post hoc explanation techniques and hide the underlying biases of $f$.
\spara{Input} The adversary provides the following to our framework: 1)~the biased classifier $f$ which they intend to deploy in the real world and, 2) an input dataset $\mathcal{X}$ that is sampled from the real world input data distribution $\mathcal{X}_{dist}$ on which $f$ will be applied. Note that neither our framework nor the adversary has access to $\mathcal{X}_{dist}$.
\spara{Output} The output of our framework will be a scaffolded classifier $e$ (referred to as \emph{adversarial classifier} henceforth) that behaves exactly like $f$ when making predictions on instances sampled from $\mathcal{X}_{dist}$, but will not reveal the underlying biases of $f$ when probed with leading post hoc explanation techniques such as LIME and SHAP.

\para{Intuition}
As discussed in the previous section, LIME and SHAP (and several other post hoc explanation techniques) explain individual predictions of a given black box model by constructing local interpretable approximations (e.g., linear models). Each such local approximation is designed to capture the behavior of the black box within the neighborhood of a given data point. These neighborhoods constitute synthetic data points generated by perturbing features of individual instances in the input data. However, instances generated using such perturbations could potentially be off-manifold or out-of-distribution (OOD)~\cite{mittelstadt2019explaining}. 

To better understand the nature of the synthetic data points generated via perturbations, we carried out the following experiment. 
First, we perturb input instances using the approach employed by LIME (See previous section). 
We then run principal component analysis (PCA) on the combined dataset containing original instances as well as the perturbed instances, and reduce the dimensionality to 2.
%
As we can see from Figure~\ref{fig:pca}, the synthetic data points generated from input perturbations are distributed significantly differently from the instances in the input data. This result indicates that detecting whether a  data point is a result of a perturbation or not is not a challenging task, and thus approaches that rely heavily on these perturbations, such as LIME, can be \emph{gamed}. 

This intuition underlies our proposed approach.
By being able to differentiate between data points coming from the input distribution and instances generated via perturbation, an adversary can create an adversarial classifier (\emph{\scaffolding}) that behaves like the original classifier (perhaps be extremely discriminatory) on the input data points, but behaves arbitrarily differently (looks unbiased and \emph{fair}) on the perturbed instances, thus effectively fooling LIME or SHAP into generating innocuous explanations.
\hide{
The aforementioned result implies: 1) post hoc explanation techniques employing input perturbations are not very reliable and, 2) it might be possible for adversarial entities to get away with deploying highly discriminatory classifiers which can fool these explanatory techniques. For instance, an adversarial actor might build a classifier that is highly discriminatory on input instances (in-sample data points) but very fair on perturbed instances (out-of-sample data points). Since approaches such as LIME rely so heavily on perturbed instances, they will output explanations which make the underlying classifier look innocuous (since the underlying model is in fact fair on perturbed instances). 
}
Next, we formalize this intuition and explain our framework for building adversarial classifiers that can fool explanation techniques. 

\para{Building Adversarial Classifiers}
Let $f$ be the biased classifier described earlier. The adversary would like to design a framework such that if and when end users generate explanations of this black box, post hoc techniques can be fooled into thinking that the model is innocuous.
%
Recall that the real world data on which this classifier is likely to be applied follows a distribution $\mathcal{X}_{dist}$, and $\mathcal{X}$ is a set of $N$ data points sampled from this distribution that the adversary has access to. To fool the post hoc techniques, the adversary could design an adversarial classifier that exhibits biased behavior on instances sampled from $\mathcal{X}_{dist}$, and remain unbiased on instances that do not come from $\mathcal{X}_{dist}$. 
Since the feature importances output by LIME and SHAP rely heavily on perturbed instances (which may typically be OOD samples, e.g. Figure~\ref{fig:pca}), the resulting explanations will make the classifier designed by the adversary look innocuous.

Assuming $\psi$ is a \emph{unbiased} classifier (e.g., makes predictions based on innocuous features that are uncorrelated with sensitive attributes), the adversarial classifier $e$ takes the following form: 
\[
    e(x)= 
\begin{cases}
    f(x),& \text{if } x \in \mathcal{X}_{dist}\\
    \psi(x),              & \text{otherwise}
\end{cases}
\]
In order to create such a classifier, we need to be able to decide whether a given data point $x$ comes from $\mathcal{X}_{dist}$ or not. 

\para{Detecting OOD Samples}
To build a classifier $\texttt{is\_OOD}$ that detects if a given data point is an \emph{out-of-distribution} (OOD) sample ($\texttt{is\_OOD}(x) = \texttt{True}$) or not, we construct a new dataset from the instances in $\mathcal{X}$. 
Specifically, we \emph{perturb} (more details in Experimental Evaluation) each instance $x$ in $\mathcal{X}$ to generate a new instance $x_p$. Let us denote the set of all the instances generated via perturbation as $\mathcal{X}_p$. The instances in $\mathcal{X}$ are then assigned the class label $\texttt{False}$ indicating that they are not OOD samples, while the instances in $\mathcal{X}_p$ are assigned the class label $\texttt{True}$ (indicating that they are OOD samples) unless they are already in $\mathcal{X}$. 
We then train an off-the-shelf classifier on the combined dataset $\mathcal{X} \cup \mathcal{X}_p$ and their corresponding class labels (assigned as discussed above).  

\section{Experimental Results}\label{sec:expts}
In this section, we discuss the detailed experimental evaluation of our framework. First, we analyze the effectiveness of the adversarial classifiers generated by our framework. More specifically, we test how well these classifiers can mask their biases by fooling multiple post hoc explanation techniques. Next, we evaluate the robustness of our adversarial classifiers by measuring how their effectiveness varies with changes to different parameters (e.g., weighting kernel, background distribution). 
Lastly, we present examples of post hoc explanations (both LIME and SHAP) of individual instances in the data to demonstrate how the biases of the classifier $f$ are successfully hidden. 

\begin{table*}
    \centering
    \begin{tabular}{lp{0.5cm}p{5.05cm}ll}
    \toprule
        \bf Dataset & \bf Size & \bf Features & \bf Positive Class & \bf Sensitive Feature \\ \midrule
        \bf COMPAS & 6172 & \em criminal history, demographics, COMPAS risk score, jail and prison time &   High Risk (81.4\%) & African-American (51.4\%)\\ \addlinespace
        \bf Communities \& Crime & 1994 & \em race, age, education, police demographics, marriage status, citizenship & Violent Crime Rate (50\%)  & White Population (continuous) \\ \addlinespace
        \bf German Credit & 1000 & \em account information, credit history, loan purpose, employment, demographics & Good Customer (70\%) & Male (69\%)\\
        \bottomrule
    \end{tabular}
    \vspace{0.2in}
    \caption{Summary of Datasets}
    \vspace{-0.2in}
    \label{tab:data}
\end{table*}

\para{Datasets}
We experimented with multiple datasets pertaining to diverse yet critical real world applications such as recidivism risk prediction, violent crime prediction, and credit scoring. Below, we describe these datasets in detail (See Table~\ref{tab:data} for detailed statistics).
Our first dataset is the \textbf{COMPAS} dataset which was collected by ProPublica~\cite{compas}. This dataset captures detailed information about the criminal history, jail and prison time, demographic attributes, and COMPAS risk scores for 6172 defendants from Broward Couty, Florida.  The sensitive attribute in this dataset is race -- 51.4\% of the defendants are African-American. Each defendant in the data is labeled either as high-risk or low-risk for recidivism. 
Our second dataset is \textbf{Communities and Crime} (CC) that captures various socio-economic and law enforcement aspects, as well as crime across various communities in the US~\cite{redmond2002data}. This dataset contains information 1994 communities (each community is a data point) in total. The sensitive attribute in this dataset is the percentage of Caucasian population. This percentage varies widely across communities and ranges from 2.68\% to 99.63\%. Each community is assigned a label depending on if the violent crime proportion in that community is above the median (violent crime rampant) or not. 
The third dataset we use is the \textbf{German credit} dataset from the UCI repository~\cite{blake1999repository}.  This dataset captures financial and demographic information (including account information, credit history, employment, gender) of about 1000 loan applicants. 
The sensitive attribute in this dataset is gender -- 69\% of the individuals are male. Each individual in the dataset is labeled either as a good customer or a bad one depending on their credit risk. 

\para{Experimental Setup}
Here, we discuss the implementation details of our framework. We randomly split all the datasets into training (90\%) and test (10\%) sets. We build our framework using the training set and evaluate it on the test set.

\spara{Biased Classifier $f$} We construct $f$ to be a perfectly discriminatory classifier that uses a sensitive feature i.e., $f$ makes predictions purely based on race in the case of COMPAS and Communities and Crime datasets (if \emph{race = African American}, then predict 1, otherwise set predict 0), and based on gender for the German Credit dataset.

\spara{Perturbations \& OOD classifier} 
For LIME, we generate perturbations for every data point in the data set $\mathcal{X}$ by adding random noise sampled from $\mathcal{N}(0,1)$ to each feature value. 
For SHAP, we randomly choose a subset of features for each data point and mark their values as missing by replacing them with their corresponding values from background distribution\footnote{See \citet{lundberg17:a-unified} for more details.}.
We choose the background distribution for SHAP to be cluster centers from kmeans with 10 clusters. We use these newly generated perturbations as well as the input data points to train the OOD classifier to detect OOD samples (See Proposed Framework section). We use the standard sklearn implementation of a random forest with 100 tree estimators as our OOD classifier.

\spara{Unbiased Classifier $\psi$} We build the unbiased classifier $\psi$ by constructing synthetic \emph{uncorrelated features} that have zero correlation with sensitive attributes (e.g., race or gender). We experiment with one or two uncorrelated features. When we only have one uncorrelated feature in a particular experiment, $\psi$ solely uses that to make predictions (if uncorrelated feature = 1, then predict 1, else predict 0). On the other hand, when we have two uncorrelated features in an experiment, we  base the predictions on the xor of those two features. Note that $\psi$ does not have to be restricted to always use synthetic uncorrelated features. It can also use any other existing feature in the data to make predictions. We experiment with synthetic uncorrelated features on COMPAS and CC dataset, and with \emph{Loan Rate \% Income} feature on the German credit dataset.

\spara{Generating Explanations} We use default LIME tabular implementation without discretization, and the default Kernel SHAP implementation with kmeans with $10$ clusters as the background distribution. 

\begin{figure}[tb]
    \centering
        \includegraphics[width=1\columnwidth,clip,trim=20 0 25 0]{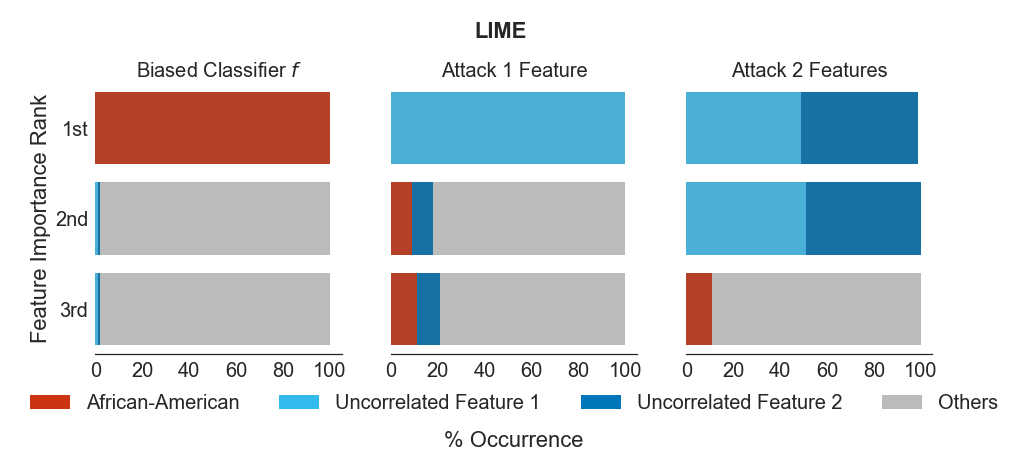}
        
        \includegraphics[width=1\columnwidth,clip,trim=20 0 25 0]{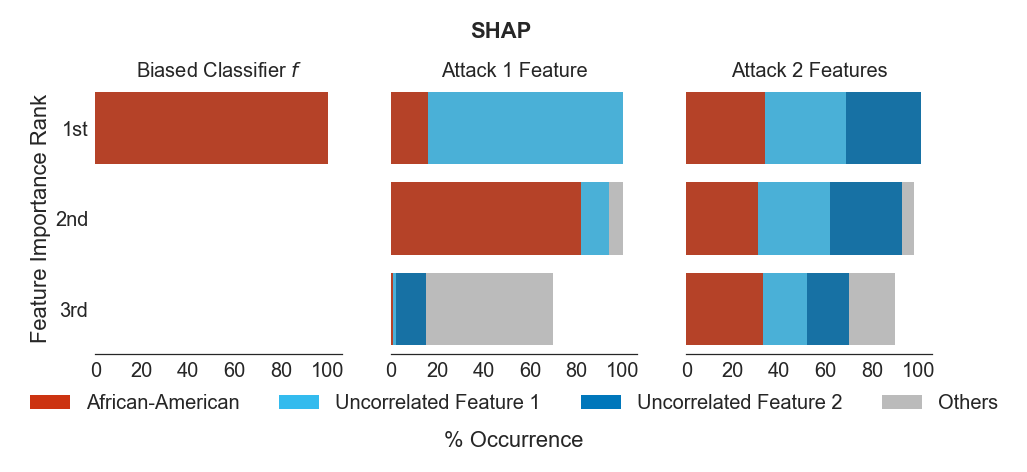}
    \vspace{-0.2in}
    \caption{\textbf{COMPAS:} \% of data points for which each feature (color coded) shows up in top 3 (according to LIME and SHAP's ranking of feature importances) for the biased classifier $f$ (left), our adversarial classifier where $\psi$ uses only one uncorrelated feature to make predictions (middle), and our adversarial classifier where $\psi$ uses two uncorrelated features to make predictions (right). 
    }
    \label{fig:compas_graph}
    \vspace{-0.2in}
\end{figure}

\begin{figure}[tb]
    \centering
    \includegraphics[width=1\columnwidth,clip,trim=20 0 25 0]{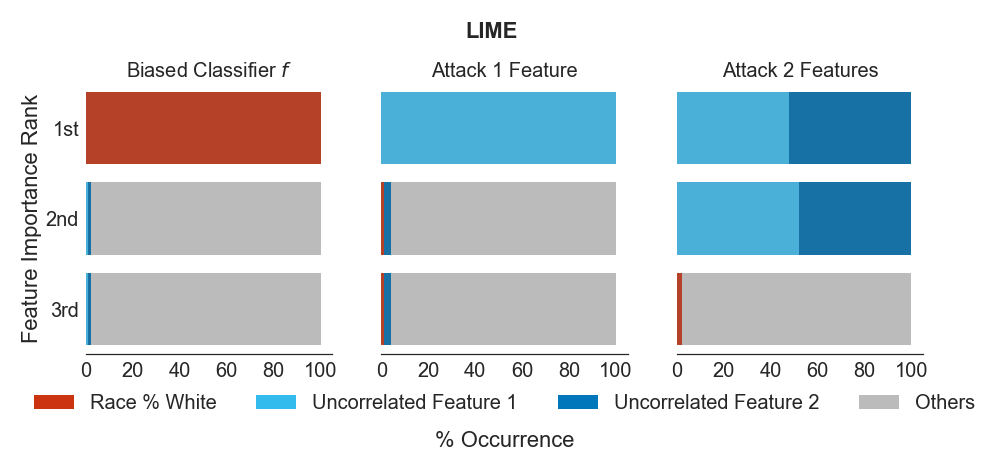}
    \vspace{0.1in}
     \includegraphics[width=1\columnwidth,clip,trim=20 0 25 0]{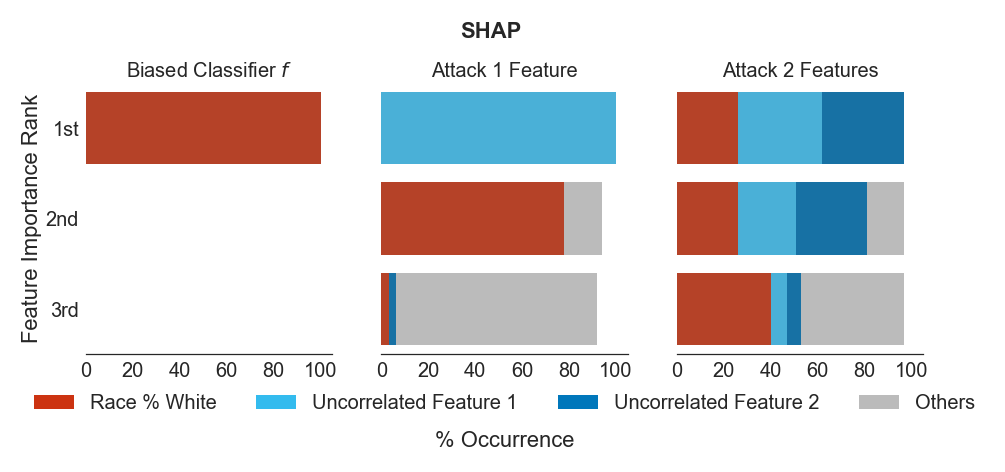}
    \caption{\textbf{Communities and Crime:} Similar to Fig~\ref{fig:compas_graph}; \emph{Race \% White} is the sensitive feature here.}
    \label{fig:cc_graph}
\end{figure}

\begin{figure}[tb]
    \centering
    \includegraphics[width=1\columnwidth,clip,trim=20 10 25 0]{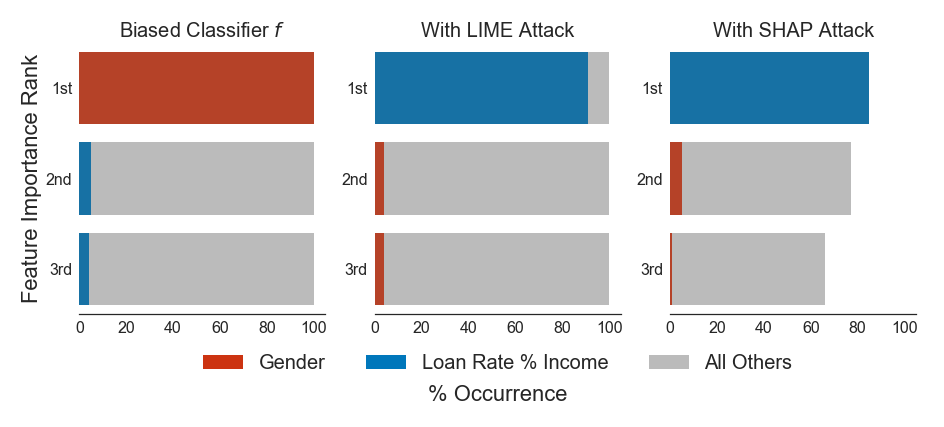}
    \caption{\textbf{German credit:} Similar to Fig~\ref{fig:compas_graph} and \ref{fig:cc_graph}, but unbiased classifier $\psi$ uses an existing feature (\emph{Loan Rate \% Income}) to make predictions, and \emph{Gender} is the sensitive feature. Feature importances for the biased classifier $f$ shown in the figure (left) are generated using LIME; SHAP also produces similar feature importance values.
    }
    \label{fig:german_graph}
\end{figure}
\subsection{Effectiveness of Adversarial Classifiers}
To evaluate how successful our \emph{attacks} are on LIME and SHAP, we compute the percentage of data points for which race, uncorrelated features (in case of COMPAS and CC) or \emph{Loan Rate \% Income} (in case of German credit data) show up in top 3 when features are ranked based on feature attributions output by LIME and SHAP. In figures~\ref{fig:compas_graph}, \ref{fig:cc_graph}, and \ref{fig:german_graph}, we show the results of these experiments.\footnote{Code available at: \url{https://github.com/dylan-slack/Fooling-LIME-SHAP}}

In case of \textbf{LIME}, 
when a single feature is used for the attack i.e., $\psi$ uses a single feature for making predictions, the adversarial classifier $e$ successfully shifts the feature importance
in LIME from the true informative feature (the sensitive feature used by $f$) to the uncorrelated feature in 100\% of held-out points for COMPAS (Figure~\ref{fig:compas_graph} - top and middle) and CC (Figure~\ref{fig:cc_graph} - top and middle), and to the \emph{Loan \% Income} feature in 91\% for German (Figure~\ref{fig:german_graph} - middle). 
Furthermore, the sensitive feature appears at 2nd or 3rd rank position roughly only as often as other features. 
When both uncorrelated features are used to define $\psi$ (for COMPAS and CC), the 1st and 2nd rank positions for nearly all the data points are occupied by these features, with the sensitive feature entering into 3rd place at about the same rate as other features (Figures~\ref{fig:compas_graph} \& ~\ref{fig:cc_graph} - top right). 
These results demonstrate that the LIME explanation technique has been effectively fooled by the adversarial classifier $e$ for all three datasets. 

For \textbf{SHAP}, when a single uncorrelated feature is used for the attack, the adversarial classifier $e$ successfully shifts the feature importance
from the sensitive feature in 100\% of data points in CC (Figure~\ref{fig:cc_graph} - bottom and middle), 84\% of data points in COMPAS (Figure~\ref{fig:compas_graph} - bottom and middle), and 85\% of data points in German. 
When two uncorrelated features are used in the attack (for COMPAS and CC), the adversarial classifier is less successful in removing the bias feature from 1st place in the ranking, succeeding in only 67\% and 71\% instances for COMPAS and CC respectively (Figures~\ref{fig:compas_graph} \& ~\ref{fig:cc_graph} - bottom right). 
This is due to SHAP's local accuracy property that ensures that feature attributions must add up to the difference between a given prediction and the average prediction for the background distribution. This property will tend to distribute feature attributions across several features when it is not possible to identify a single most informative feature. 
\begin{figure}
    \centering
    \begin{subfigure}{\columnwidth}
    \centering
    \includegraphics[width=0.85\columnwidth]{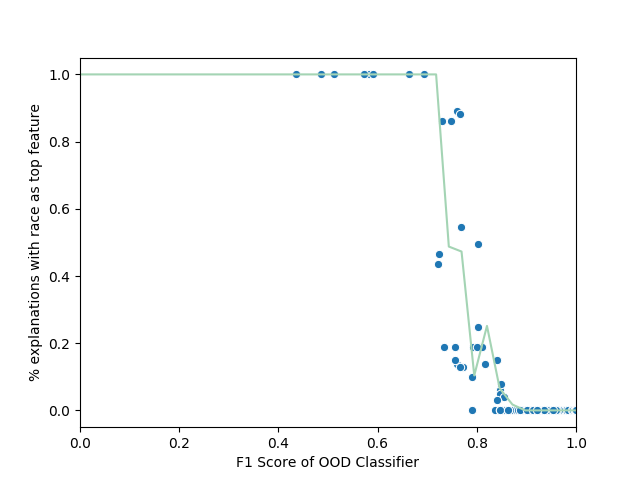}
    \caption{LIME COMPAS Sensitivity Analysis}
    \end{subfigure}
    
    \begin{subfigure}{\columnwidth}
    \centering
    \includegraphics[width=0.85\columnwidth]{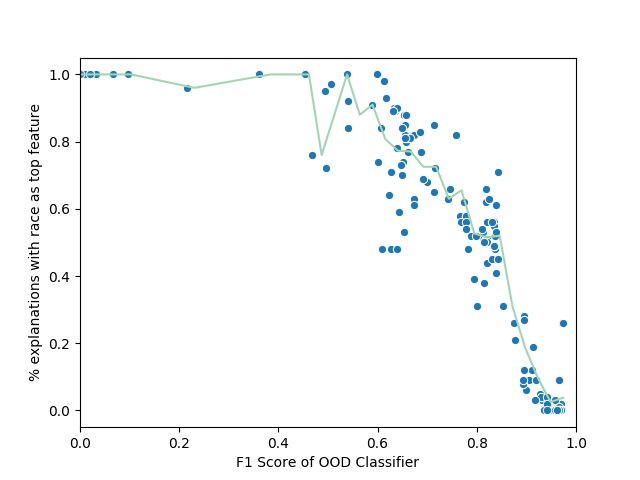}
    \caption{SHAP COMPAS Sensitivity Analysis}
    \end{subfigure}
    \caption{Effectiveness of our attack: \% of data points for which race is the most important feature vs. the F1 score of the OOD classifier (perturbation detection) for both LIME and SHAP on the COMPAS dataset over a variety of models (fit with line plot over mean of binned points). 
    }
    \label{fig:pert_detect}
\end{figure}

To ensure that the adversarial classifiers $e$ are faithfully approximating the original classifiers $f$ on instances from the original dataset, we evaluate the \emph{fidelity} of $e$ w.r.t. $f$ 
by computing the percentage of held-out instances on which their predictions agree.
In case of LIME, our adversarial classifiers perfectly reproduce the classifier $f$ on all datasets 100\% of the time. On the other hand, in case of SHAP, our adversarial classifiers are slightly less effective: 75\% for CC, 80\% for German, and 91\% for COMPAS. These results indicate that our approach has constructed adversarial classifiers $e$ that successfully fool LIME and SHAP, making them look like the unbiased $\psi$ classifiers, while behaving like the original biased classifiers $f$ on input data. \vspace{-0.1in}

\subsection{Effect of Perturbation Detection Accuracy}
Our approach relies on the OOD classifier to detect whether a given instance is a result of perturbation or not. 
Here, we investigate how accurate does this classifier have to be for our attacks to be effective. 
We train a number of different OOD classifiers with varying accuracies (F1 scores) on the COMPAS dataset. 
For each OOD classifier, we evaluate the effectiveness of the resulting attacks (measured as percentage of data points for which race is the most important feature in the post hoc explanations).
The results are presented in Figure~\ref{fig:pert_detect}.  
When the OOD classifiers are unable to adequately differentiate between perturbed instances and points from the input dataset, 
both LIME and SHAP explanations detect race as the most important feature i.e., the adversarial attacks are not successful.
However, once the OOD classifier is more accurate, the adversarial attacks are successful, and both LIME and SHAP do not identify race as the most important feature.
Between the two, LIME requires the OOD classifier to be more accurate for the attack to be effective, but is very vulnerable to these attacks once the F1 score is above 0.8. 
SHAP, on the other hand, starts responding to the attacks with a less accurate OOD classifier (F1 score $\sim$ 0.45), however, the effect is more gradual; it needs a very accurate OOD classifier to completely hide the bias. 
These results indicate that our attacks are effective as long as it is possible to differentiate between perturbed instances and input data points with a reasonable accuracy.

\begin{table}
        \centering
        \begin{tabular}{lrr}
            \toprule
             \bf Feature & \bf Before & \bf After attack  \\
             \midrule
             Race & -0.42 & 0.0 \\
              Uncorrelated feature 1 & 0.0 & 0.42 \\
             Average of other features & 0.0 & 0.0 \\
    
             \bottomrule
        
        \end{tabular}
        \vspace{0.2in}
        \caption{Feature coefficients of LIME explanations for an instance from COMPAS, before and after an attack ($\psi$ uses a single feature).}
        \label{tab:lime_example}
        \vspace{-0.5in}
    \end{table}

    \begin{figure}
    
    \begin{subfigure}{\columnwidth}
    \includegraphics[width=1\columnwidth]{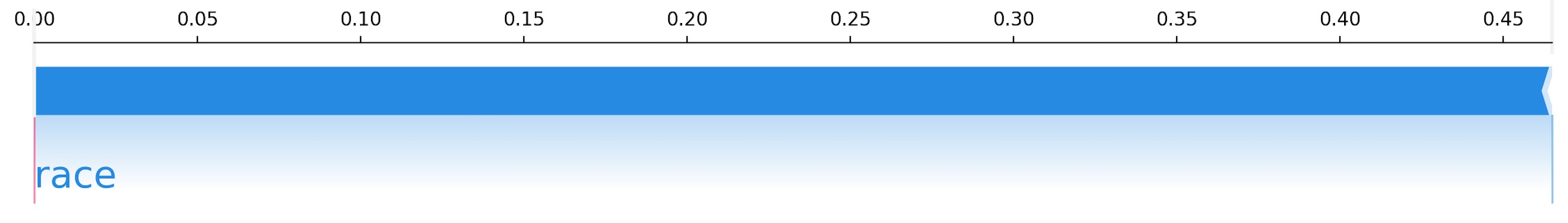}
    \caption{Example SHAP explanation for biased classifier $f$}
    \end{subfigure}
    
    \centering
    \begin{subfigure}{\columnwidth}
    \includegraphics[width=1\columnwidth]{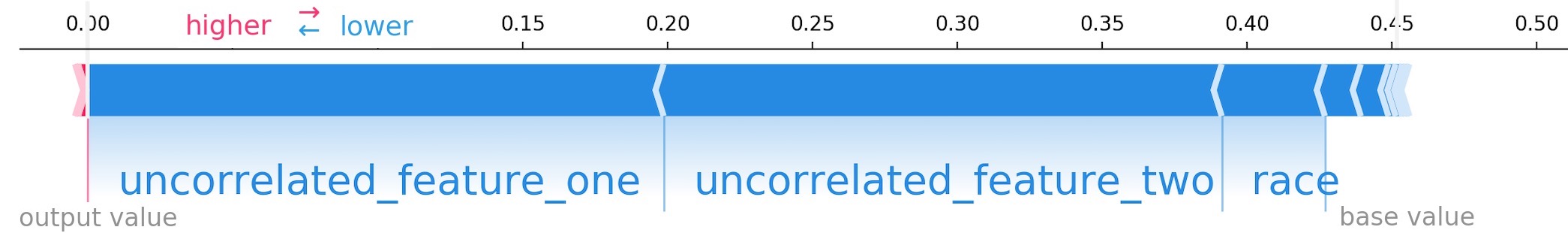}
    \caption{SHAP explanation of the same instance after the attack ($e$)}
    \end{subfigure}
     \caption{Example SHAP explanations for a data point with biased classifier $f$ (top) and adversarial classifier $e$ (bot.).}
    \label{fig:shap_example}
    \vspace{-0.1in}
    \end{figure}

\subsection{Robustness to Hyperparameters}
We analyze the behavior of our framework when different parameters are varied \emph{at test time} when generating LIME and SHAP explanations for COMPAS; we summarize the key results here. 
First, we observe that classifiers that are adversarial to LIME are ineffective against SHAP explanations, while adversarial classifiers designed for SHAP have varying effectiveness against LIME. 
On varying the LIME kernel width between $0.15 \sqrt{M}$ to $0.95 \sqrt{M}$ (default kernel width is $0.75 \sqrt{M}$, where $M$ is the number of features), the attacks remain equally effective. 
Additionally, changing the LIME kernel to use $l1$ norm for distance instead of the default $l2$ norm also does not impact the effectiveness. 
For SHAP, we train our adversarial classifier assuming the default background distribution of K-means with 10 clusters;
5 and 15 clusters at test-time are also equally effective.
Using a background distribution of all zeroes for SHAP makes our attacks even more pronounced. 

\subsection{Example Explanations} Here, we present few example explanations from the COMPAS dataset that demonstrate the effectiveness of the technique. 
In case of LIME explanations in Table~\ref{tab:lime_example}, we see that \emph{Race} as a feature is completely hidden from the explanation after the attack, even though it was the sole important feature for the original classifier $f$.
Similarly, for SHAP explanations in Figure~\ref{fig:shap_example}, the sole important feature (\emph{race}) is considerably hidden in the explanation after the attack, although not completely nullified as in the LIME explanation. 

\section{Related Work} \label{sec:related}
This work lies at the intersection of various emerging sub fields of machine learning such as post hoc explanation methods and bias detection. Below, we provide a brief overview of the related work.

\para{Perturbation-based Explanation Methods}
Perturbation-based methods are a popular tool for post hoc feature attribution. In addition to LIME~\cite{ribeiro16:kdd} 
and SHAP~\cite{lundberg17:a-unified},
a number of other perturbation-based techniques have been proposed in literature. For instance, anchors \cite{ribeiro2018anchors} use  (non-linear) rules to express more actionable local explanations. GAM~\cite{ibrahim2019global} interprets local attributions as conjoined weighted rankings and uses k-medoids clustering to identify prototypical explanations. 




\para{Criticism of Post hoc Explanations}
\citet{rudin2019stop} argues that post hoc explanations are not reliable, as these explanations are not necessarily faithful to the underlying models and present correlations rather than information about the original computation.
\citet{ghorbani2019interpretation} 
show that some explanation techniques can be highly sensitive to small perturbations in the input even though the underlying classifier's predictions remain unchanged. 
\citet{mittelstadt2019explaining} note that perturbation points created in LIME and SHAP are not at all intuitive, especially in case of structured data.

\para{Adversarial Explanations}
There has been some recent research on manipulating explanations in the context of image classification.  \citet{dombrowski2019explanations} show that by modifying inputs in a way that is imperceptible to humans, 
they can arbitrarily change saliency maps. 
\citet{heo2019fooling} also propose similar attacks on saliency maps. 

\para{Interpretability and Bias Detection}
\citet{doshi2017towards} argue that interpretability can help us evaluate if a model is biased or discriminatory. On the other hand, \citet{lipton2016mythos} posits that post hoc explanations can never definitively prove or disprove unfairness of any given classifier. \citet{selbst2018intuitive} and \citet{kroll2016accountable} show that even if a model is completely transparent, it is hard to detect and prevent bias due to the existence of correlated variables. More recently, \citet{aivodji2019fairwashing} demonstrated that post hoc explanations can potentially be exploited to \emph{fairwash} i.e., rationalize decisions made by unfair black-box models.

\section{Conclusions and Future Work}
We proposed a novel framework that can effectively hide discriminatory biases of any black box classifier. Our approach exploits the fact that post hoc explanation techniques such as LIME and SHAP are perturbation-based to create a \emph{\scaffolding} around the biased classifier such that its predictions on input data distribution remain biased, but its behavior on the perturbed data points is controlled to make the post hoc explanations look completely innocuous.  Extensive experimentation with real world data from criminal justice and credit scoring domains demonstrates that our approach is effective at generating adversarial classifiers that can fool post hoc explanation techniques, finding that LIME is more vulnerable than SHAP. 
Our findings thus suggest that existing post hoc explanation techniques are not sufficient for ascertaining discriminatory behavior of classifiers in sensitive applications. 

This work paves way for several interesting future research directions in ML explainability. First, it would be interesting to systematically study if other classes of post hoc explanation techniques (e.g., gradient based approaches) are also vulnerable to adversarial attacks. Second, it would be interesting to develop new techniques for building \emph{adversarially robust} explanations that can withstand attacks such as the ones outlined in this work.


\section{Acknowledgements}

We would like to thank the anonymous reviewers for their feedback, and Scott Lundberg for insightful discussions.
This work is supported in part by the Allen Institute for Artificial Intelligence (AI2), NSF award \#IIS-1756023, and Google. The views expressed are those of the authors and do not reflect the official policy or position of the funding agencies.

%
\bibliographystyle{ACM-Reference-Format}
\bibliography{main,sameer}
\end{document}